\documentclass{article}

\usepackage{microtype}
\usepackage{graphicx}
\usepackage{subcaption}
\usepackage{booktabs} % for professional tables

\usepackage{hyperref}

\usepackage[preprint]{icml2026}

\usepackage{amsmath}
\usepackage{amssymb}
\usepackage{mathtools}
\usepackage{amsthm}

\usepackage[utf8]{inputenc} % allow utf-8 input
\usepackage[T1]{fontenc}    % use 8-bit T1 fonts
\usepackage{hyperref}       % hyperlinks
\usepackage{url}            % simple URL typesetting
\usepackage{booktabs}       % professional-quality tables
\usepackage{amsfonts}       % blackboard math symbols
\usepackage{nicefrac}       % compact symbols for 1/2, etc.
\usepackage{microtype}      % microtypography
\usepackage{xcolor}         % colors
\usepackage{graphicx}       % for figures
\usepackage{algorithm}      % for algorithms
\usepackage{algorithmic}    % for algorithms
\usepackage{subcaption}     % for subfigures
\usepackage{wrapfig}
\usepackage{authblk}

\usepackage[capitalize,noabbrev]{cleveref}

\theoremstyle{plain}

\theoremstyle{definition}

\theoremstyle{remark}

\usepackage[textsize=tiny]{todonotes}

\icmltitlerunning{Credit Assignment via Neural Manifold Noise Correlation}

\begin{document}

\twocolumn[
  \icmltitle{Credit Assignment via Neural Manifold Noise Correlation}

  % It is OKAY to include author information, even for blind submissions: the
  % style file will automatically remove it for you unless you've provided
  % the [accepted] option to the icml2026 package.

  % List of affiliations: The first argument should be a (short) identifier you
  % will use later to specify author affiliations Academic affiliations
  % should list Department, University, City, Region, Country Industry
  % affiliations should list Company, City, Region, Country

  % You can specify symbols, otherwise they are numbered in order. Ideally, you
  % should not use this facility. Affiliations will be numbered in order of
  % appearance and this is the preferred way.
  \icmlsetsymbol{equal}{*}

  \begin{icmlauthorlist}
    \icmlauthor{Byungwoo Kang}{1,2,3}
    \icmlauthor{Maceo Richards}{3}
    \icmlauthor{Bernardo Sabatini}{1,2,3}
  \end{icmlauthorlist}

  \icmlaffiliation{1}{Howard Hughes Medical Institute}
  \icmlaffiliation{2}{Department of Neurobiology, Harvard Medical School}
  \icmlaffiliation{3}{Kempner Institute for the Study of Natural and Artificial Intelligence, Harvard University}

  \icmlcorrespondingauthor{Byungwoo Kang}{byungwoo\_kang@hms.harvard.edu}
  \icmlcorrespondingauthor{Bernardo Sabatini}{bernardo\_sabatini@hms.harvard.edu}

  % You may provide any keywords that you find helpful for describing your
  % paper; these are used to populate the "keywords" metadata in the PDF but
  % will not be shown in the document
  \icmlkeywords{Neuroscience, Credit assignment problem}

  \vskip 0.3in
]

% this must go after the closing bracket ] following \twocolumn[ ...

% This command actually creates the footnote in the first column listing the
% affiliations and the copyright notice. The command takes one argument, which
% is text to display at the start of the footnote. The \icmlEqualContribution
% command is standard text for equal contribution. Remove it (just {}) if you
% do not need this facility.

% Use ONE of the following lines. DO NOT remove the command.
% If you have no special notice, KEEP empty braces:
\printAffiliationsAndNotice{}  % no special notice (required even if empty)
% Or, if applicable, use the standard equal contribution text:
% \printAffiliationsAndNotice{\icmlEqualContribution}

\begin{abstract}
Credit assignment—how changes in individual neurons and synapses affect a
network’s output—is central to learning in brains and machines. Noise correlation, which estimates gradients by correlating perturbations of activity with changes in output, provides a biologically plausible solution to credit assignment but scales poorly as accurately estimating the  Jacobian requires that the number of perturbations scale with network size. Moreover, isotropic noise conflicts with neurobiological observations that neural activity lies on a low-dimensional manifold. To address these drawbacks, we propose \emph{neural manifold noise correlation} (NMNC), which performs credit assignment using perturbations restricted to the neural manifold. We show theoretically and empirically that the Jacobian row space aligns with the neural manifold in trained networks, and that manifold dimensionality scales slowly with network size. NMNC substantially improves performance and sample efficiency over vanilla noise correlation in convolutional networks trained on CIFAR-10, ImageNet-scale models, and recurrent networks. NMNC also yields representations more similar to the primate visual system than vanilla noise correlation. These findings offer a mechanistic hypothesis for how biological circuits could support credit assignment, and suggest that biologically inspired constraints may enable, rather than limit, effective learning at scale.
\end{abstract}

\section{Introduction}

The credit assignment problem—determining how individual neurons and synapses contribute to a network's output—is fundamental to learning in both artificial and biological neural networks \citep{minsky_steps_1961,rumelhart_learning_1986}. Backpropagation solves this elegantly but requires biologically implausible features: symmetric forward and backward weights, distinct forward and backward  passes, and segregation of forward and backward pass activations \citep{crick_recent_1989,grossberg_competitive_1987}. At its essence, credit assignment requires estimating the Jacobian of the network, the gradients of the network output with respect to hidden unit activations.

Noise correlation methods estimate gradients by injecting noise and correlating it with output changes \cite{williams_simple_1992,werfel_learning_2003}. Unlike feedback alignment approaches \cite{lillicrap_random_2016,nokland_direct_2016}, noise correlation directly approximates forward-pass gradients with local learning rules. The simplest form, weight perturbation, perturbs individual synaptic weights and observes changes in output  \cite{jabri_weight_1992}. A more efficient variant, node perturbation,
perturbs neural activities rather than individual weights \cite{williams_simple_1992,fiete_gradient_2006}. This approach underlies many modern attempts to develop biologically plausible learning rules \cite{bartunov_assessing_2018,kunin_two_2020,meulemans_credit_2021,meulemans_minimizing_2022}. However, the number of perturbations required to accurated estiamte the Jacobian scales with network size \cite{werfel_learning_2003, ren_scaling_2023}. In addition, isotropic noise conflicts with empirical evidence that neural activity—not only task-related activity, but also its trial-to-trial variability and spontaneous activity—lies on a low-dimensional manifold often referred to as \emph{neural manifold} \cite{cunningham_dimensionality_2014,huang_circuit_2019,lin_nature_2015,gardner_toroidal_2022,chaudhuri_intrinsic_2019,dimakou_predictive_2025,luczak_default_2012,luczak_spontaneous_2009,kenet_spontaneously_2003,engel_new_2019}.

In this work, we connect these two observations and examine if \emph{we can exploit the structure of the neural manifold to make noise correlation scalable.} We provide theoretical and empirical evidence that the gradients lie approximately within the same low-dimensional manifold as the activity itself. In addition, we show that manifold dimensionality scales slowly with network size. Based on these motivations, we propose \emph{neural manifold noise correlation} (NMNC), which estimates a neural manifold online, injects noise along the manifold directions, and correlates output fluctuations with the low-dimensional noise. We evaluate NMNC on deep convolutional networks, ImageNet-scale models, and recurrent neural networks, demonstrating substantial improvements over vanilla noise correlation in performance and sample efficiency. Finally, we show that training convolutional networks with NMNC yields more primate-like visual representations than with vanilla noise correlation.

\section{Background and motivation}

\subsection{Alignment of Jacobian row space and neural manifold}
\label{subsec:alignment}

\begin{figure*}[t]
\centering
\includegraphics[width=\linewidth]{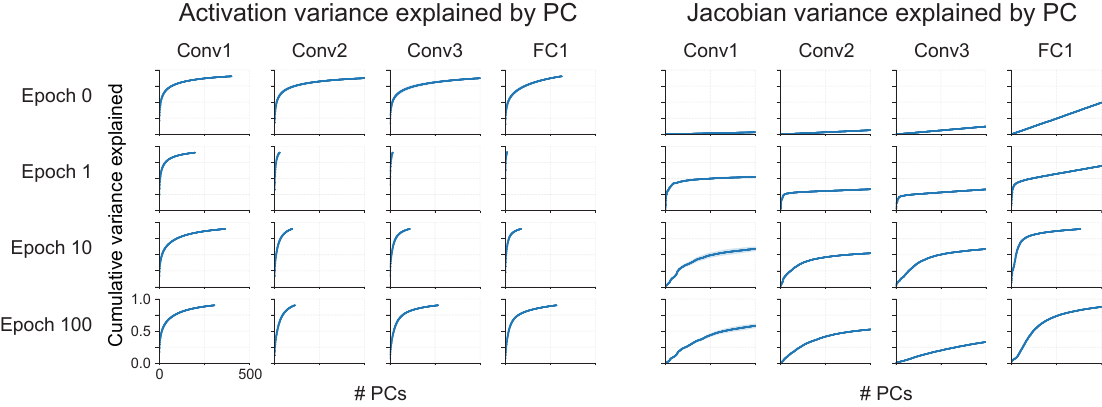}
\caption{(Left) Variance of activations explained by principal components across training epochs for each layer of convolutional neural networks trained on CIFAR-10. Epoch 0 refers to the network prior to training, and Epoch 100 is the last epoch. (Right) Same analysis applied to the Jacobian variance. Curves are shown up to 90\% cumulative variance explained. mean $\pm$ std, $n=5$ seeds.}
\label{fig:1}
\end{figure*}

Consider the Jacobian $J_l = \frac{\partial \mathbf{y}}{\partial \mathbf{x}_l} \in \mathbb{R}^{n_o \times n_l}$, which maps perturbations in layer $l$'s activation space to changes in the network output y. Let $\mathcal{M}_l \subset \mathbb{R}^{n_l}$ denote the neural manifold at layer $l$, and let $U_l \in \mathbb{R}^{n_l \times d_l}$ be an orthonormal basis for the subspace spanned by $\mathcal{M}_l$. Any perturbation decomposes into components parallel and orthogonal to the manifold:
\begin{equation}
    \boldsymbol{\xi} = U_l U_l^T \boldsymbol{\xi} + (I - U_l U_l^T) \boldsymbol{\xi} = \boldsymbol{\xi}_{\parallel} + \boldsymbol{\xi}_{\perp}
\end{equation}

The central point is that the network's downstream layers have been trained exclusively on activations drawn from $\mathcal{M}_l$. Consequently, the network's response to $\boldsymbol{\xi}_{\perp}$ is essentially undefined and driven by the random structure at initialization. In contrast, $\boldsymbol{\xi}_{\parallel}$ probes the part of $J_l$ learned during training that captures the meaningful input-output relationships.

A complementary view comes from tracking how gradients shape downstream weights. Note that
\begin{equation}
    J_l = \frac{\partial \mathbf{y}}{\partial \mathbf{x}_l}=\frac{\partial \mathbf{y}}{\partial \mathbf{s}_{l+1}}\frac{\partial \mathbf{s}_{l+1}}{\partial \mathbf{x}_l}=\frac{\partial \mathbf{y}}{\partial \mathbf{s}_{l+1}}W_{l+1},
\label{eq:J_l_W_l+1}
\end{equation}
where $\mathbf{s}_{l+1} = W_{l+1} \mathbf{x}_{l}$ and $\mathbf{x}_l=\phi(\mathbf{s}_l)$.
Because gradient descent updates $W_{l+1}$ with $\Delta W_{l+1} \propto \boldsymbol{\delta}_{l+1}x_l^T$, the row space of the learned part of $W_{l+1}$ is spanned by the presynaptic activations $x_l^T$. Thus, via \eqref{eq:J_l_W_l+1}, it follows that the row space of $J_l$ is spanned by the history of the $x_l^T$, up to the random initial component. \footnote{A similar observation has previously been made in \cite{singhal_how_2023}.} Since the rows of the Jacobian approximately lie in the neural manifold, we only need to probe directions within the manifold to estimate it well enough.

We empirically confirmed the above theoretical considerations in a convolutional neural network trained by backpropagation on CIFAR-10 (\cref{fig:1}) (see \Cref{app:architectures} for architecture details). With training, the Jacobian aligns with the neural manifold (defined by PCA), leading to significant fractions of its variance explained by a relatively small number of principal components (PCs).\footnote{We note that substantially more PCs are required to capture the variance of the Jacobian than that of the activations. This gap arises for at least two reasons. First, although the learned part of the Jacobian is spanned by the history of activations during training, PCA in this analysis is performed on the current activations only. Second, the Jacobian is also determined by downstream weights, which are in turn determined by the history of downstream activations and error signals. This effectively changes the relative importance of different PCs and makes low-variance PCs account for a substantial fraction of the Jacobian variance, much more than they do for the activation variance.}

\subsection{Scaling of neural manifold dimensionality with the network size}

% \begin{wrapfigure}{r}{0.5\linewidth}
% \centering
%   \includegraphics[width=\linewidth]{Fig2.pdf}
%   \caption{Network size vs. neural manifold dimensionality (TwoNN or \#PCs for 90\% variance). mean $\pm$ std, $n=5$ seeds.}
%   \label{fig:2}
% \end{wrapfigure}
\begin{figure}
\centering
  \includegraphics[width=\linewidth]{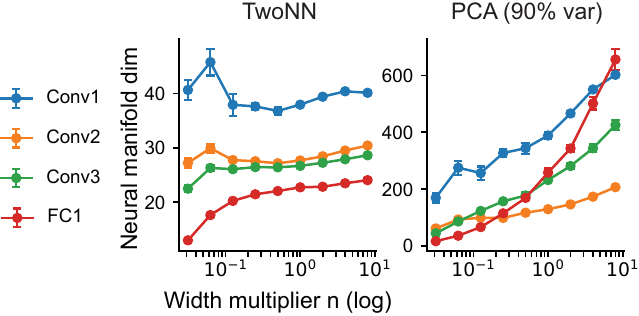}
  \caption{Network size vs. neural manifold dimensionality (TwoNN or \#PCs for 90\% variance). mean $\pm$ std, $n=5$ seeds.}
  \label{fig:2}
\end{figure}

If the relevant manifold dimension $d_l$ remains small as networks scale, then restricting perturbations to this subspace can improve the sample efficiency of noise correlation. We therefore varied network width over two orders of magnitude(see \Cref{app:intrinsic_dim} for details), holding architecture and dataset fixed, and estimated intrinsic dimensionality using TwoNN \cite{facco_estimating_2017,sharma_scaling_2022} (see \Cref{app:twonn} for a self-contained explanation) and PCA. Across layers, manifold dimensionality grows slowly with width and remains far below $n_l$ (\Cref{fig:2}), suggesting that the \emph{effective} dimensionality of credit assignment can be much smaller than the raw activation dimension especially in large neural networks.

\section{Neural manifold noise correlation}
Based on the above theoretical and empirical motivations, we propose that credit assignment can be performed using \textit{neural manifold noise correlation} (NMNC). NMNC learns feedback weights by performing noise correlation \emph{within} each layer's activity manifold.
For each layer $l$, we maintain (i) a low-dimensional basis $U_l \in \mathbb{R}^{n_l\times d_l}$ (estimated online from activations) and (ii) feedback weights $B_l \in \mathbb{R}^{n_l \times n_o}$.

\paragraph{Feedback learning (noise correlation).}
Every $b$ training iterations, we update $U_l$ (using incremental PCA \cite{ross_incremental_2008}), sample $\boldsymbol{\zeta}_l \sim \mathcal{N}(0,I_{d_l})$, form a manifold-restricted perturbation $\boldsymbol{\xi}_l = U_l\boldsymbol{\zeta}_l$, and run a noisy forward pass to obtain $\Delta \mathbf{y} = \tilde{\mathbf{y}}-\mathbf{y}$. We then update the feedback weights with an exponential moving average:
\begin{equation}
B_l \leftarrow (1-\eta_B)B_l + \eta_B\cdot \frac{1}{N_b}\left(\boldsymbol{\xi}_l \Delta\mathbf{y}^T\right),
\end{equation}
where $N_b$ is batch size. Full pseudocode is given in \Cref{alg:nmnc} (\Cref{app:nmnc_pseudocode}).

\paragraph{Forward-weight updates.}
Given the current $B_l$, we compute a pseudo-error at each layer from the output error $\boldsymbol{\delta}_{\mathrm{out}}$:
\begin{equation}
\boldsymbol{\delta}_l = \phi'(\mathbf{s}_l)\odot (B_l \boldsymbol{\delta}_{\mathrm{out}}),
\end{equation}
and update forward weights locally via $\Delta W_l = -\eta\,\boldsymbol{\delta}_l \mathbf{x}_{l-1}^T$.

\paragraph{Key differences from vanilla noise correlation.}
Vanilla noise correlation (VNC) samples isotropic noise $\boldsymbol{\xi}_l \sim \mathcal{N}(0, \sigma^2 I_{n_l})$ in the full activation space. NMNC samples noise in the low-dimensional manifold ($d_l$ dimensions) and then projects it to the full space via $U_l$. This reduces variance and improves sample efficiency. To ensure fair comparison, we match noise magnitudes in NMNC and VNC: $\sigma_{\text{VNC}} = \sqrt{d_l / n_l}\sigma_{\text{NMNC}}$. 

\section{Experiments}

We evaluate NMNC as a scalable, perturbation-based credit
assignment mechanism across three regimes: (i) direct-feedback learning in a convolutional network
trained on CIFAR-10, (ii) ImageNet-scale training using AlexNet with layerwise feedback (Weight
Mirror), and (iii) recurrent networks trained via weight perturbation. Unless otherwise stated, all
methods use identical optimizers and hyperparameters for the forward weights, and the output layer
is always trained using exact gradients (see \Cref{app:cifar10_arch} for
architectures and \Cref{app:cifar10_training} for training details).

\subsection{Performance and sample efficiency of NMNC}

\begin{figure*}[h]
\centering
\includegraphics[width=\linewidth]{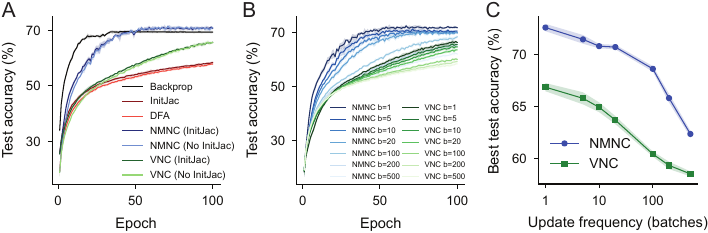}
\caption{Performance and sample efficiency of NMNC and VNC on CIFAR-10. (A) Test accuracy vs.\ epochs for different learning rules. (B) Test accuracy vs.\ epochs for varying
  frequencies of feedback update for NMNC and VNC (No InitJac). Feedback weights are updated every $b$ batch. (C) Best test accuracy vs.\ noise correlation frequency. Same data as (B). mean $\pm$ std, $n=5$ seeds.}
\label{fig:3}
\end{figure*}

\paragraph{Setup.}
We train the CIFAR-10 convolutional network described in \Cref{app:cifar10_arch} using (i) standard
backpropagation, (ii) direct feedback alignment (DFA; fixed random feedback), (iii) an ``InitJac''
baseline with fixed feedback weights set to the \emph{initial} Jacobian (evaluated on random Gaussian inputs), and (iv) learned-feedback
variants using VNC or NMNC.
For NMNC, the neural manifold basis at each hidden layer is estimated online via incremental PCA, and perturbations are restricted to that subspace.

\paragraph{Main comparison.}
\Cref{fig:3}A shows that NMNC substantially outperforms VNC and DFA and
approaches backpropagation performance. This improvement is obtained under a
fixed perturbation budget: both VNC and NMNC learn feedback weights via noise correlation, but
NMNC concentrates perturbations along directions that are most functionally relevant for the
trained network (as motivated in \Cref{fig:1}).

\paragraph{InitJac vs. No InitJac}
Although InitJac variants show early advantages, learned-feedback methods (NMNC/VNC) and DFA close this gap as training progresses. We therefore default to the No-InitJac setting for the remainder of the study, isolating the effect
of how perturbations are sampled (full-space versus manifold-restricted).

\paragraph{Sample efficiency vs. feedback-update frequency.}
Increasing the update interval $b$ reduces the number of perturbation samples used to learn the feedback weights.
\Cref{fig:3}B--C
shows that NMNC maintains higher accuracy than VNC across a wide range of $b$, consistent with reducing the effective dimensionality of the estimation problem from $n_l$ to
$d_l \ll n_l$.

\subsection{Mechanisms underlying NMNC's advantage over VNC}

\begin{figure*}[h]
\centering
\includegraphics[width=\linewidth]{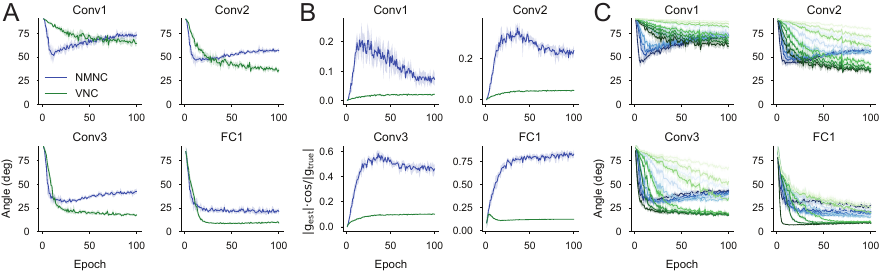}
\caption{Alignment between true and estimated gradients in activation space (see \Cref{fig:8} for alignment in weight space). (A) Cosine similarity angle between the true and estimated gradients for NMNC and VNC across layers. (B) Normalized magnitude of the estimated gradient projected onto the true gradient direction for NMNC and VNC across layers. (C) Same as (A) but for varying frequencies of feedback update. The color scheme is the same as in \Cref{fig:3}B. mean $\pm$ std, $n=5$ seeds.}
\label{fig:4}
\end{figure*}

To better understand why NMNC improves learning, we compare the \emph{pseudo-gradients} induced by
learned feedback weights to the true backpropagation gradients (computed only for purpose of their analysis; see Appendix~E for details). Across layers, we find two consistent effects:

\paragraph{Mechanism 1: improved early gradient alignment.}
\Cref{fig:4}A and C show that NMNC yields better alignment to the true gradient (smaller angle) early in
training and in lower layers, which have higher-dimensional activity space. In some settings, VNC can match or slightly exceed NMNC later in training when feedback
updates are frequent and enough samples accumulate. However, probably because alignment is more important earlier in learning, NMNC's early alignment advantage translates into better performance, even when VNC catches up later.

\paragraph{Mechanism 2: larger effective step along the true gradient.}
We also quantify the magnitude of the pseudo-gradient component that lies along the true gradient
direction. \Cref{fig:4}B shows that NMNC produces a larger projected pseudo-gradient across layers,
meaning that NMNC typically takes a larger effective step in the direction that decreases the loss.
Interestingly, we observe regimes in which VNC attains higher cosine alignment late in training even though
NMNC still has a larger projected magnitude; we provide an explanation below. \footnote{Given that NMNC can produce pseudo-gradients with larger norm than VNC, one might ask
whether VNC could compensate simply by increasing the forward learning rate. Empirically, we found
that larger learning rates for VNC often destabilize training, consistent with additional stochasticity
from SGD and sample-to-sample variability in gradient estimation.}

\paragraph{Relationship between pseudo- and true gradient.}
For small perturbations $\Delta y \approx J_l \xi$, the expected feedback-weight
update in \Cref{alg:nmnc} is
\begin{equation}
\mathbb{E}[\Delta B_l]
= \mathbb{E}[\xi\Delta y^\top ]
\approx  \mathbb{E}[\xi\xi^\top]J_l^\top
= \Sigma_lJ_l^\top,
\label{eq:delta_B}
\end{equation}

where $\Sigma_l := \mathbb{E}[\xi\xi^\top]$ is the noise covariance. Ignoring the slow drift
of $J_l$, the feedback weights converge to $B_l^\star \propto \Sigma_l J_l^\top $ (see \Cref{alg:nmnc}), and the
pseudo-gradient is
\begin{equation}
\tilde g_l
= B_l^{\star}\delta_{\text{out}}
\approx \Sigma_l J_l^\top \delta_{\text{out}}
= \Sigma_l g_l.
\label{eq:pseudo_g}
\end{equation}
Thus noise correlation returns the true gradient pre-multiplied by the noise covariance.

\paragraph{An explanation for early alignment.}
Let $g_l$ denote the true backprop gradient with respect to the activations of layer $l$, and let $\xi_l$ be the noise injected into that layer when learning the feedback weights.
After $k$ noise-correlation updates, the resulting pseudo-gradient can be written as
\begin{equation}
\tilde g_l^{(k)} = \hat\Sigma_l^{(k)} g_l,
\qquad
\hat\Sigma_l^{(k)} = \frac{1}{k}\sum_{i=1}^k \xi_l^{(i)}\xi_l^{(i)\top},
\end{equation}
where $\hat\Sigma_l^{(k)}$ is the empirical noise covariance. Its expectation is the true covariance
$\Sigma_l = \mathbb{E}[\xi_l\xi_l^\top]$, so we may decompose
\begin{equation}
\tilde g_l^{(k)} = \Sigma_l g_l + \eta_l^{(k)},
\qquad
\eta_l^{(k)} := (\hat\Sigma_l^{(k)} - \Sigma_l)g_l.
\end{equation}

For Gaussian noise one can show (using standard fourth-moment identities) that
\begin{equation}
\mathbb{E}\bigl[\|\eta_l^{(k)}\|^2\bigr]
= \frac{1}{k}\Bigl[
\mathrm{tr}(\Sigma_l)\, g_l^\top\Sigma_l g_l + g_l^\top \Sigma_l^2 g_l
\Bigr].
\label{eq:*}
\end{equation}

In VNC the noise is isotropic in the full $n_l$-dimensional activation space,
$\Sigma_l^{\mathrm{VNC}} = \tfrac{\tau_l}{n_l}I_{n_l}$, whereas in NMNC it is restricted to the
$d_l$-dimensional neural manifold with projector $P_l := U_l U_l^\top$,
$\Sigma_l^{\mathrm{NMNC}} = \tfrac{\tau_l}{d_l}P_l$, with $\tau_l = \mathbb{E}\|\xi_l\|^2$ matched
between methods. Writing
\begin{equation}
\alpha_l := \frac{\|P_l g_l\|^2}{\|g_l\|^2} \in [0,1]
\end{equation}

for the fraction of gradient energy lying in the manifold, and using \eqref{eq:*}, we obtain
\begin{equation}
\mathbb{E}\|\eta_l^{(k)}\|^2_{\mathrm{VNC}} \approx \frac{\tau_l^2}{k n_l},
\qquad
\mathbb{E}\|\eta_l^{(k)}\|^2_{\mathrm{NMNC}} \approx \frac{\tau_l^2\alpha_l}{k d_l}.
\end{equation}

Approximating $\eta_l^{(k)}$ as noise uncorrelated with signal $\Sigma_l g_l$ and replacing the norm of $\eta_l^{(k)}$ in
the denominator by its expectation yields the following expressions for the expected squared cosine
between the pseudo-gradient and the true gradient:
% \begin{equation}
%     \mathbb{E}\!\left[\cos^2\bigl(\tilde g_l^{(k)}, g_l\bigr)\right]_{\mathrm{VNC}}
% \approx \frac{k}{k + n_l + 1},
% \qquad
% \mathbb{E}\!\left[\cos^2\bigl(\tilde g_l^{(k)}, g_l\bigr)\right]_{\mathrm{NMNC}}
% \approx \frac{\alpha_l k}{k + d_l + 1}.
% \label{eq:**}
% \end{equation}

\begin{equation}
\begin{aligned}
\mathbb{E}\!\left[\cos^2\bigl(\tilde g_l^{(k)}, g_l\bigr)\right]_{\mathrm{VNC}}
&\approx \frac{k}{k + n_l + 1}, \\
\mathbb{E}\!\left[\cos^2\bigl(\tilde g_l^{(k)}, g_l\bigr)\right]_{\mathrm{NMNC}}
&\approx \frac{\alpha_l k}{k + d_l + 1}.
\end{aligned}
\label{eq:**}
\end{equation}

For small $k$, these scale as $\mathbb{E}[\cos^2]_{\mathrm{VNC}} \approx k/n_l$ and
$\mathbb{E}[\cos^2]_{\mathrm{NMNC}} \approx \alpha_l k/d_l$, so NMNC has better early alignment
whenever $\alpha_l > d_l/n_l$. In our CIFAR-10 setting, $d_l/n_l$ is small while the Jacobian row
space is strongly aligned with the activity manifold (\Cref{fig:1}), making this condition easy to
satisfy. Intuitively, NMNC only needs to estimate a $d_l$-dimensional preconditioner that captures
most of the gradient energy, whereas VNC must estimate an $n_l$-dimensional object from the same
number of samples.

As $k\to\infty$, \eqref{eq:**} predicts $\cos^2_{\mathrm{VNC}}\to 1$ while $\cos^2_{\mathrm{NMNC}}\to \alpha_l$,
so VNC can eventually achieve slightly higher cosine alignment than NMNC if it accumulates many
perturbation samples. In practice, with feedback updates every $b$ batches, each layer only sees
$k\approx T/b$ samples over $T$ training iterations. When $T/b \ll n_l$ (e.g.\ larger $b$), VNC
never reaches its asymptotic regime and NMNC maintains higher alignment throughout training, as
observed in \Cref{fig:4}C.

\paragraph{An explanation for the pseudo-gradient magnitude and projection.}
Noise correlation learns feedback weights proportional to $J_l\Sigma_l$, and therefore returns a
pseudo-gradient proportional to $\Sigma_l g_l$. Under matched perturbation energy
$\tau_l=\mathrm{tr}(\Sigma_l)$, VNC yields $\tilde g_l^{\mathrm{VNC}} = \tfrac{\tau_l}{n_l}g_l$ whereas
NMNC yields $\tilde g_l^{\mathrm{NMNC}} = \tfrac{\tau_l}{d_l}P_l g_l$. Their squared norms satisfy
\begin{equation}
    \frac{\|\tilde g^{\mathrm{NMNC}}_l\|}{\|\tilde g^{\mathrm{VNC}}_l\|}
=
\frac{n_l}{d_l}
\frac{\|P_l g_l\|}{\|g_l\|}.
\label{eq:***}
\end{equation}

Thus, whenever $\|P_l g_l\|/\|g_l\| > d_l/n_l$ (the same condition as above), NMNC produces a
pseudo-gradient with larger expected norm.
Moreover, the component along the true gradient direction scales as
\[
\frac{\big\|\mathrm{Proj}_{g_l}(\tilde g^{\mathrm{NMNC}}_l)\big\|}
     {\big\|\mathrm{Proj}_{g_l}(\tilde g^{\mathrm{VNC}}_l)\big\|}
=
\frac{n_l}{d_l}\frac{\|P_l g_l\|^2}{\|g_l\|^2},
\]
so NMNC can take a larger effective step along the true gradient even when its cosine alignment is
slightly lower. This provides a parsimonious explanation for why, late in training, VNC can
sometimes show higher alignment while NMNC continues to exhibit a larger projected pseudo-gradient
(\Cref{fig:4}B) and better learning.

\subsection{Application of NMNC to ImageNet-scale models}

% \begin{wrapfigure}{r}{0.6\linewidth}
% \centering
%   \includegraphics[width=\linewidth]{Fig5.pdf}
%   \caption{Comparison of NMNC and VNC on ImageNet. Test accuracy of AlexNet on ImageNet when
%   trained with (i) Backprop, (ii) the weight mirror algorithm using vanilla noise correlation (VNC),
%   and (iii) the weight mirror algorithm using neural manifold noise correlation (NMNC). mean $\pm$
%   std, $n=5$ seeds.}
%   \label{fig:5}
% \end{wrapfigure}
\begin{figure}{h}
\centering
  \includegraphics[width=\linewidth]{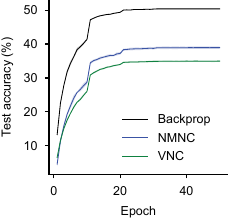}
  \caption{Comparison of NMNC and VNC on ImageNet. Test accuracy of AlexNet on ImageNet when
  trained with (i) Backprop, (ii) the weight mirror algorithm using vanilla noise correlation (VNC),
  and (iii) the weight mirror algorithm using neural manifold noise correlation (NMNC). mean $\pm$
  std, $n=5$ seeds.}
  \label{fig:5}
\end{figure}

Having established NMNC as a practical learning rule on CIFAR-10, we next test whether it can be
applied at ImageNet scale. This regime is also relevant from a neuroscience perspective: training on
large and diverse natural image datasets is associated with the emergence of more primate-like visual
representations in deep networks \cite{conwell_large-scale_2024}.

\paragraph{Why AlexNet and why layerwise feedback.}
We use AlexNet (rather than more recent architectures with very large activation tensors) because online
incremental PCA becomes prohibitively expensive in our current implementation. Direct feedback from
the output layer (as in the CIFAR-10 experiments) was unstable or substantially degraded, suggesting
that a single linear map from output error to early-layer activations is a poor approximation at this scale.
We therefore adopt a \emph{layerwise} feedback scheme based on the Weight Mirror approach
\cite{akrout_deep_2019}, and compare isotropic perturbations (VNC) to
manifold-restricted perturbations (NMNC) (see \Cref{app:imagenet_arch,app:imagenet_training} for details).

\paragraph{Results.}
Training with NMNC significantly outperforms VNC on ImageNet, although a gap remains compared to backpropagation (\Cref{fig:5}). This indicates that restricting perturbations to the neural manifold remains beneficial at this scale. We also observe that in this layerwise -feedback regime, VNC can exhibit higher cosine alignment than NMNC (\Cref{fig:10}A): the feedback objects being learned (i.e. transposed kernels) are substantially lower-dimensional than the direct-feedback matrices used for CIFAR-10, and we update feedback weights
every batch to achieve reasonable performance. Despite this, NMNC yields larger pseudo-gradient magnitudes and projected steps, consistent with its improved accuracy (\Cref{fig:10}B).

\subsection{Neural representations of ImageNet-scale models trained with Backprop, NMNC and VNC}

\begin{figure*}[h]
\centering
\includegraphics[width=\linewidth]{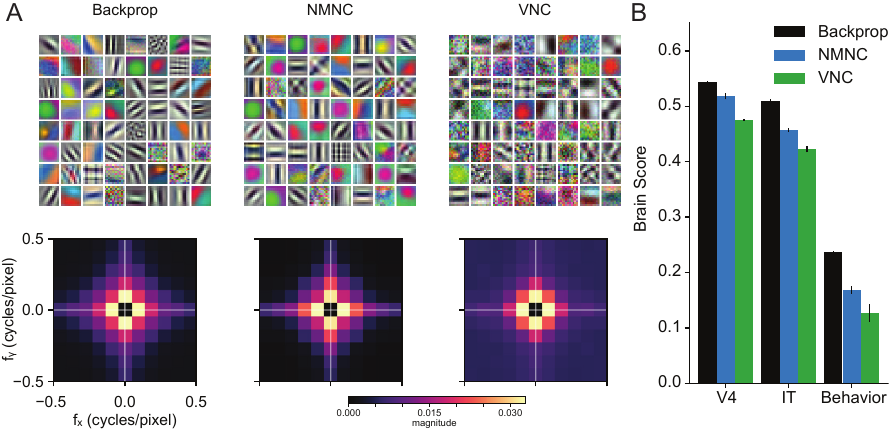}
\caption{Comparison of neural representations emerging from NMNC and VNC. (A) (Top row) Conv1 kernels from models trained with Backprop, NMNC and VNC. (Bottom row) Discrete Fourier transform on the Conv1 kernels. mean, n = 5 seeds. (B) V4, IT, and Behavior Brain Scores for models trained with Backprop, NMNC and VNC. mean ± std, n = 5 seeds.}
\label{fig:6}
\end{figure*}

Beyond task performance, we examined whether restricting perturbations to the neural manifold influences
the \emph{representations} that emerge during learning. We analyzed the ImageNet-scale models
trained with backpropagation, NMNC, and VNC.

\paragraph{First-layer filters.}
A classic qualitative signature of ImageNet-trained AlexNet is that first-layer convolutional kernels
resemble Gabor-like filters, reminiscent of V1 receptive fields \cite{krizhevsky_imagenet_2012}. \Cref{fig:6}A shows that all three
learning rules indeed broadly produce Gabor-like kernels. However, VNC-trained models additionally
exhibit prominent salt-and-pepper, high-frequency patterns superimposed on these filters. This is
reflected in the Fourier-domain visualization in \Cref{fig:6}A (bottom row), showing that VNC yields kernels with stronger high-frequency components.

\paragraph{Brain-score evaluation.}
To more systematically compare representations to the primate ventral visual stream, we evaluated
the trained models using Brain Score metrics for V4, IT, and behavior \cite{schrimpf_brain-score_2018,schrimpf_integrative_2020}. \Cref{fig:6}B shows that
backpropagation yields the highest Brain Scores overall, followed by NMNC and then VNC. A
higher Brain Score does not by itself imply that a learning rule is biologically correct, but the consistent
ordering and the qualitative filter differences suggest that incorporating structure in perturbations
(i.e.\ aligning them with natural activity patterns) can bias learning toward more brain-like
representations.

\subsection{Application of NMNC to recurrent neural networks}

\begin{figure*}[h]
\centering
\includegraphics[width=\linewidth]{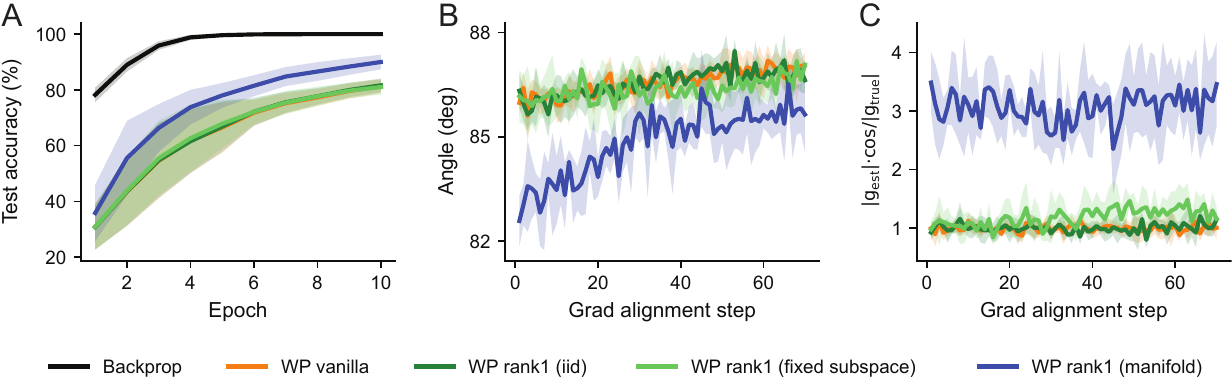}
\caption{Neural manifold noise correlation in recurrent networks.
(A) Performance of RNNs trained with Backprop and variants of weight perturbation (WP) on a sequential memory task. ``WP vanilla'' refers to standard full-rank WP with i.i.d.\ perturbations. ``WP rank1 (iid)'' uses rank-1 perturbations with i.i.d.\ factors. ``WP rank1 (fixed subspace)'' samples rank-1 perturbations from a fixed random subspace with dimensionality matched to the neural manifold. ``WP rank1 (manifold)'' samples rank-1 perturbations from the neural manifold of the hidden state. (B) Cosine similarity angle between the true and estimated gradient for $W_{hh}$. (C) Normalized magnitude of the estimated gradient projected onto the true gradient direction for $W_{hh}$. mean $\pm$ std, $n=5$ seeds.}
\label{fig:7}
\end{figure*}

Biological circuits are highly recurrent, and
credit assignment in recurrent neural networks (RNNs) poses an additional challenge: backpropagation
through time (BPTT) requires transporting error signals \emph{back in time} to the same neurons
rather than to upstream layers. A naive application of node-perturbation-style noise correlation to
RNNs would still require learning feedback pathways that deliver appropriate temporal credit.

\paragraph{Manifold-structured low-rank weight perturbation.}
In the RNN setting, weight perturbation (WP) has a unique advantage: it does not require an explicit
temporal feedback pathway, because it directly perturbs the recurrent weights and correlates the
resulting loss change with the perturbation. Motivated by recent work on low-rank perturbation
schemes for large-scale optimization \cite{sarkar_evolution_2025}, we tested whether \emph{restricting}
the perturbations to the neural manifold of the hidden state can improve WP in RNNs.\footnote{Low-rank
WP does not remove the fundamental variance scaling of standard WP with the number of parameters; its
main advantage is computational/hardware efficiency \cite{sarkar_evolution_2025}. See \Cref{app:lowrank_wp_variance} for an explanation. Here we isolate a
complementary effect: choosing the perturbation subspace to match the network's activity manifold.}

Specifically, we compared: (i) standard full-rank WP with i.i.d.\ perturbations, (ii) rank-1 WP with
i.i.d.\ factors, (iii) rank-1 WP whose factors lie in a fixed random subspace matched in dimension to
the neural manifold, and (iv) rank-1 WP whose factors are sampled from the \emph{neural manifold} of
the hidden state (estimated online). For fair comparison, all perturbation types were scaled to have matched magnitude (see \Cref{app:rnn_training} for details of architecture and training).

\paragraph{Results.}
On a sequential memory task, rank-1 manifold WP achieves the best performance among WP
variants (\Cref{fig:7}A). Consistent with the feedforward results, manifold-structured perturbations
also yield better gradient estimates: rank-1 manifold WP exhibits higher gradient alignment
(\Cref{fig:7}B) and larger projected pseudo-gradient magnitude (\Cref{fig:7}C) for the recurrent weight
matrix $W_{hh}$. These results suggest that the core NMNC principle of constraining perturbations to
the neural manifold is applicable beyond feedforward networks and can improve perturbation-based learning in recurrent settings as well.

\section{Discussion}

Noise correlation offers an appealing route to biologically plausible credit assignment because it can estimate gradients using only forward computations and locally available correlations.
However, its classic formulations using isotropic noise scale poorly as the number of samples required to obtain useful feedback signals scales with the dimensionality of the activity space.

\paragraph{Neural manifolds reduce the effective dimensionality of credit assignment}
Our central observation is that trained networks (and biological circuits) tend to operate on
low-dimensional activity manifolds, and that the functionally relevant components of the Jacobian
become aligned with these manifolds through learning (\Cref{fig:1}).
NMNC exploits this structure by performing noise correlation \emph{on the manifold} rather than in
the full activity space. This reduces the effective dimension of the estimation problem and improves
sample efficiency. Empirically, NMNC closes much of the gap between perturbation-based learning and
backpropagation on CIFAR-10 (\Cref{fig:3}), provides a consistent advantage over vanilla noise
correlation at ImageNet scale when combined with layerwise feedback learning (\Cref{fig:5}), and
extends to recurrent networks via manifold-structured low-rank weight perturbation (\Cref{fig:7}).

% \paragraph{Mechanistic interpretation: covariance-shaped feedback in a low-dimensional subspace.}
% Noise correlation does not learn the Jacobian directly; in expectation it learns feedback weights of the
% form $B_l^\star \propto \Sigma_l J_l^\top$, where $\Sigma_l$ is the covariance of the injected perturbations
% (\Cref{eq:delta_B}). Consequently, applying the learned feedback to the output error yields
% $\tilde g_l \approx \Sigma_l g_l$ (\Cref{eq:pseudo_g}): the true gradient passed through a linear operator
% set by how variability is injected. NMNC chooses $\Sigma_l$ to be supported on the activity manifold,
% so this operator concentrates its gain on the $d_l$ manifold directions that carry most of the gradient
% energy while suppressing off-manifold directions that the network rarely visits. This makes the
% feedback-estimation problem effectively $d_l$-dimensional, improving early gradient alignment, and it
% also increases the update component along $g_l$ under matched perturbation energy, explaining why
% NMNC can take larger effective steps even when cosine alignment is not uniformly higher throughout
% training (\Cref{fig:4}).

\paragraph{Implications for neuroscience and biologically plausible learning.}
From a biological standpoint, NMNC suggests a concrete hypothesis: correlated variability aligned
with a circuit's intrinsic activity manifold is not merely ``noise,'' but may provide the structured
perturbations needed for credit assignment when combined with global output or performance signals (e.g.\
neuromodulators). In this picture, the brain may not need to inject independent perturbations across
all neurons. Instead, it can exploit the low-dimensional structure of population activity to learn
effective credit assignment with far fewer degrees of freedom.

\paragraph{Limitations and future directions.}
First, our current implementation uses PCA, which provides only a linear subspace approximation to
potentially nonlinear manifolds. Incorporating nonlinear manifold models (e.g.\ learned encoders or locally linear subspaces) could further reduce bias while preserving sample efficiency. Second, online manifold estimation can be computationally expensive for modern architectures with very large activations; developing more efficient and
hardware-friendly estimators would improve practical scalability. Third, the brain likely generates
manifold-aligned fluctuations via mechanisms different from our specific implementation involving incremental PCA.
Encouragingly, there are biologically plausible proposals for online PCA and related dimensionality
reduction algorithms \cite{qiu_neural_2012,oja_simplified_1982,oja_neural_1989,oja_principal_1992,kung_neural_1990,sanger_optimal_1989,foldiak_adaptive_1989,linsker_improved_2005,minden_biologically_2018,pehlevan_hebbiananti-hebbian_2015}. Incorporating such mechanisms into NMNC would
make the algorithm more biologically realistic. Fourth, the remaining gap on performance on ImageNet indicates that complementary improvements in feedback parameterizations, initialization, and additional local learning signals are needed to close the gap with backpropagation. Fifth, while noise correlation methods including NMNC address the problem of symmetric forward and backward weights, one of the key biological implausibilities of backpropagation, they do not by themselves resolve the other problems mentioned in the introduction. Notably, several recent proposals that solve these additional issues also rely on noise correlation to learn feedback pathways, and we expect NMNC to similarly improve their sample efficiency relative to VNC \cite{meulemans_credit_2021,meulemans_minimizing_2022,meulemans_least-control_2022}.

More broadly, our results indicate that incorporating biologically inspired structure---rather than
treating it as a constraint---can enable effective learning at scale. We hope NMNC motivates further
work connecting low-dimensional population dynamics, structured variability, and plausible credit
assignment mechanisms in both artificial and biological neural systems.

% % Acknowledgements should only appear in the accepted version.
% \section*{Acknowledgements}
% This work has been made possible in part by a gift from the Chan Zuckerberg Initiative Foundation to establish the Kempner Institute for the Study of Natural and Artificial Intelligence at Harvard University.

% \textbf{Do not} include acknowledgements in the initial version of the paper
% submitted for blind review.

% If a paper is accepted, the final camera-ready version can (and usually should)
% include acknowledgements.  Such acknowledgements should be placed at the end of
% the section, in an unnumbered section that does not count towards the paper
% page limit. Typically, this will include thanks to reviewers who gave useful
% comments, to colleagues who contributed to the ideas, and to funding agencies
% and corporate sponsors that provided financial support.

\section*{Impact Statement}
This work aims to advance understanding of how biologically plausible learning rules could support credit assignment in high-dimensional neural systems. By connecting perturbation-based learning to low-dimensional neural activity manifolds, it offers a conceptual bridge between empirical observations in neuroscience and learning algorithms studied in machine learning. The expected impact is primarily scientific, providing a framework and hypotheses that may guide future theoretical, computational, and experimental work at the interface of these fields. We do not anticipate direct societal risks specific to this contribution beyond those generally associated with progress in machine learning research.

\bibliography{zotero}
\bibliographystyle{icml2026}

%%%%%%%%%%%%%%%%%%%%%%%%%%%%%%%%%%%%%%%%%%%%%%%%%%%%%%%%%%%%%%%%%%%%%%%%%%%%%%%
%%%%%%%%%%%%%%%%%%%%%%%%%%%%%%%%%%%%%%%%%%%%%%%%%%%%%%%%%%%%%%%%%%%%%%%%%%%%%%%
% APPENDIX
%%%%%%%%%%%%%%%%%%%%%%%%%%%%%%%%%%%%%%%%%%%%%%%%%%%%%%%%%%%%%%%%%%%%%%%%%%%%%%%
%%%%%%%%%%%%%%%%%%%%%%%%%%%%%%%%%%%%%%%%%%%%%%%%%%%%%%%%%%%%%%%%%%%%%%%%%%%%%%%
\newpage
\appendix
\onecolumn

\section{NMNC pseudocode}
\label{app:nmnc_pseudocode}

\begin{algorithm}
\caption{Neural Manifold Noise Correlation (NMNC)}
\label{alg:nmnc}
\begin{algorithmic}[1]
\STATE \textbf{Input:} Network with layers $1, \ldots, L$; manifold dimensions $\{d_l\}$; PCA update interval $b$; feedback learning rate $\eta_B$; batch size $N_b$
\STATE Initialize $\{U_l\}$ and $\{B_l\}$ randomly
% Initialize $\{B_l\}$ to initial Jacobian (InitJac) or random (No InitJac)

% \STATE Initialize $\{U_l\}$ randomly; Initialize $\{B_l\}$ to initial Jacobian (InitJac) or random (No InitJac)
% \STATE \textit{// InitJac: Compute $B_l^{(0)} = \frac{\partial \mathbf{y}}{\partial \mathbf{x}_l}$ averaged over a small batch of random inputs}
% \STATE \textit{// No InitJac: Random permutation of InitJac elements (preserves statistics, destroys structure)}
\FOR{each training iteration $t$}
    \STATE Forward pass: compute activations $\{\mathbf{x}_l\}$ and output $\mathbf{y}$
    \STATE \textit{// Learn feedback weights via noise correlation:}
    \IF{$t \mod b = 0$}
        \STATE Update $\{U_l\}$ via incremental PCA on $\{\mathbf{x}_l\}$
        \FOR{each layer $l$}
            \STATE Sample low-dim noise: $\boldsymbol{\zeta}_l \sim \mathcal{N}(0, I_{d_l})$
            \STATE Project to activation space: $\boldsymbol{\xi}_l = U_l \boldsymbol{\zeta}_l$
        \ENDFOR
        \STATE Forward pass with noise: inject noise $\boldsymbol{\xi}_l$ to each layer and compute noisy output $\tilde{\mathbf{y}}$
        \STATE Compute output change: $\Delta \mathbf{y} = \tilde{\mathbf{y}} - \mathbf{y}$
        \FOR{each layer $l$}
            \STATE Update: $B_l \leftarrow (1-\eta_B) B_l + \eta_B \cdot \frac{1}{N_b}(\boldsymbol{\xi}_l \Delta \mathbf{y}^T)$
        \ENDFOR
    \ENDIF
    \STATE \textit{// Compute weight updates using learned feedback:}
    \STATE Compute output error: $\boldsymbol{\delta}_{\text{out}} = \frac{1}{N_b}(\text{softmax}(\mathbf{y}) - \mathbf{y}_{\text{target}}$)
    \FOR{each layer $l = L, \ldots, 1$}
        \STATE Compute pseudo-error: $\boldsymbol{\delta}_l = \phi'(\mathbf{s}_l) \odot (B_l \boldsymbol{\delta}_{\text{out}})$
        \STATE Update weights: $\Delta W_l = -\eta \cdot \boldsymbol{\delta}_l \mathbf{x}_{l-1}^T$
    \ENDFOR
\ENDFOR
\end{algorithmic}
\end{algorithm}

\section{Network Architectures}
\label{app:architectures}

\subsection{CIFAR-10 Architecture}
\label{app:cifar10_arch}

For CIFAR-10 experiments, we use the same convolutional network architecture used in \cite{bartunov_assessing_2018}:

\begin{table}[h]
\centering
\caption{CIFAR-10 network architecture. All convolutional and fully-connected layers (except the output layer) are followed by ReLU activations.}
\label{tab:cifar10_arch}
\begin{tabular}{@{}llllll@{}}
\toprule
Layer & Type & Input $\rightarrow$ Output & Kernel & Stride & Padding \\
\midrule
conv1 & Conv2d & $3 \times 32 \times 32 \rightarrow 64 \times 16 \times 16$ & $5 \times 5$ & 2 & 2 \\
conv2 & Conv2d & $64 \times 16 \times 16 \rightarrow 128 \times 8 \times 8$ & $5 \times 5$ & 2 & 2 \\
conv3 & Conv2d & $128 \times 8 \times 8 \rightarrow 256 \times 4 \times 4$ & $3 \times 3$ & 2 & 1 \\
fc1 & Linear & $4096 \rightarrow 1024$ & -- & -- & -- \\
fc2 & Linear & $1024 \rightarrow 10$ & -- & -- & -- \\
\bottomrule
\end{tabular}
\end{table}

The post-activation shapes and corresponding flat dimensions for each layer are:

\begin{table}[h]
\centering
\caption{Post-activation dimensions for each layer in the CIFAR-10 network.}
\label{tab:cifar10_dims}
\begin{tabular}{@{}llll@{}}
\toprule
Layer & Post-Activation Shape & Flat Dimension $n_l$ & Default \# PCs $d_l$ \\
\midrule
conv1 & $(64, 16, 16)$ & 16,384 & 512 \\
conv2 & $(128, 8, 8)$ & 8,192 & 512 \\
conv3 & $(256, 4, 4)$ & 4,096 & 512 \\
fc1 & $(1024,)$ & 1,024 & 128 \\
\bottomrule
\end{tabular}
\end{table}

% For the CIFAR-10 network,
% we use manifold dimensions $d_l = (512, 512, 512, 128)$ for (Conv1, Conv2, Conv3, FC1)
% respectively.

\subsection{ImageNet Architecture}
\label{app:imagenet_arch}

For ImageNet experiments, we use a standard AlexNet architecture with the following specification:

\begin{table}[h]
\centering
\caption{ImageNet (AlexNet) network architecture. MaxPool layers follow conv1, conv2, and conv5. Dropout ($p=0.5$) is applied before fc1 and fc2. The output layer (fc3) is always trained with exact gradients.}
\label{tab:imagenet_arch}
\begin{tabular}{@{}llllll@{}}
\toprule
Layer & Type & Input $\rightarrow$ Output & Kernel & Stride & Padding \\
\midrule
conv1 & Conv2d & $3 \rightarrow 64$ & $11 \times 11$ & 4 & 2 \\
pool1 & MaxPool2d & -- & $3 \times 3$ & 2 & 0 \\
conv2 & Conv2d & $64 \rightarrow 192$ & $5 \times 5$ & 1 & 2 \\
pool2 & MaxPool2d & -- & $3 \times 3$ & 2 & 0 \\
conv3 & Conv2d & $192 \rightarrow 384$ & $3 \times 3$ & 1 & 1 \\
conv4 & Conv2d & $384 \rightarrow 256$ & $3 \times 3$ & 1 & 1 \\
conv5 & Conv2d & $256 \rightarrow 256$ & $3 \times 3$ & 1 & 1 \\
pool5 & MaxPool2d & -- & $3 \times 3$ & 2 & 0 \\
avgpool & AdaptiveAvgPool2d & Output: $6 \times 6$ & -- & -- & -- \\
dropout1 & Dropout & $p = 0.5$ & -- & -- & -- \\
fc1 & Linear & $9216 \rightarrow 4096$ & -- & -- & -- \\
dropout2 & Dropout & $p = 0.5$ & -- & -- & -- \\
fc2 & Linear & $4096 \rightarrow 4096$ & -- & -- & -- \\
fc3 & Linear & $4096 \rightarrow 1000$ & -- & -- & -- \\
\bottomrule
\end{tabular}
\end{table}

The post-activation shapes (before pooling where applicable) and corresponding flat dimensions are:

\begin{table}[h]
\centering
\caption{Post-activation dimensions for each layer in the ImageNet network. Shapes shown are after ReLU but before any subsequent pooling operation. For noise correlation methods, we inject noise at these 7 post-ReLU locations (conv1--5, fc1--2).}
\label{tab:imagenet_dims}
\begin{tabular}{@{}llll@{}}
\toprule
Layer & Post-Activation Shape & Flat Dimension $n_l$ & Default \# PCs $d_l$ \\
\midrule
conv1 & $(64, 55, 55)$ & 193,600 & 2,048 \\
conv2 & $(192, 27, 27)$ & 139,968 & 2,048 \\
conv3 & $(384, 13, 13)$ & 64,896 & 2,048 \\
conv4 & $(256, 13, 13)$ & 43,264 & 2,048 \\
conv5 & $(256, 13, 13)$ & 43,264 & 2,048 \\
fc1 & $(4096,)$ & 4,096 & 1,024 \\
fc2 & $(4096,)$ & 4,096 & 1,024 \\
\bottomrule
\end{tabular}
\end{table}

\paragraph{Note on noise injection.} For NMNC and VNC, noise is injected after the ReLU activation at each hidden layer listed above. The output layer is always trained with exact gradients from the cross-entropy loss.

% \subsection{RNN Architecture}
% \label{app:rnn_arch}
% Explain RNN architecture.

%%%%%%%%%%%%%%%%%%%%%%%%%%%%%%%%%%%%%%%%%%%%%%%%%%%%%%%%%%%%%
\section{Training Details}
\label{app:training}

\subsection{CIFAR-10 Training Configuration}
\label{app:cifar10_training}

All CIFAR-10 models were trained using stochastic gradient descent with momentum. Forward weights were optimized with a learning rate of 0.001, while feedback weights were trained with the same learning rate ($\eta_B = 0.001$). Momentum was set to 0.9, and models were trained for 100 epochs with a batch size of 64.

For NMNC, incremental PCA was used to estimate low-dimensional activity manifolds online. For the default configuration, PCA bases were updated every 5 batches and the number of retained principal components per layer was set to $[512, 512, 512, 128]$.

No data augmentation was applied during training beyond standard per-channel normalization. Input images were normalized using dataset-wide means of $[0.4914, 0.4822, 0.4465]$ and standard deviations of $[0.2470, 0.2435, 0.2616]$. $\sigma_{\text{NMNC}}$ was set to 1.0.

\subsection{ImageNet Training Configuration}
\label{app:imagenet_training}

For ImageNet experiments, models were trained using stochastic gradient descent with momentum following \cite{xiao_biologically-plausible_2018}. Forward weights were optimized with a learning rate of 0.005, a momentum of 0.9, a weight decay of $5 \times 10^{-4}$ and a batch size of 256 for 50 epochs. A step-based learning rate schedule was applied to the forward weights, reducing the learning rate by a factor of 0.1 every 10 epochs. Feedback weights were trained using the Weight Mirror method \cite{akrout_deep_2019}, where $B_l \leftarrow (1-\lambda_B) B_l + \eta_B \cdot \frac{1}{N_b}\delta_l\, \delta_{l+1}^T$. To achieve reasonable performance, for both NMNC and VNC, the feedback weights were updated every batch. For NMNC, incremental PCA was updated every 10 batch (this was the highest frequency of PCA update that could keep up with model training. When the frequency was higher, the activations used for PCA accumulated leading to the out-of-memory error). As noted in \cite{kunin_two_2020}, the Weight Mirror method is sensitive to $\lambda_B$ and $\eta_B$, so we performed a hyperparameter search for these parameters, separately for NMNC and VNC, using Optuna \cite{akiba_optuna_2019}. Specifically, for each method, we used a tree-structured Parzen estimator with 100 sweep trials. For NMNC, the best values were $\lambda_B=0.212,\,\eta_B=0.101$, and for VNC, $\lambda_B=0.414,\, \eta_B=0.0243$. The same optimization and scheduling settings were used across all ImageNet experiments.

\subsection{RNN Training Configuration}
\label{app:rnn_training}

We evaluated recurrent learning on a sequential memory task defined by parameters $(L,S,K)=(0,5,5)$. Each input is a sequence of integers between $0$ and $K-1$ of length $T=2S+L$. The input presents $S$ random symbols (uniformly drawn from $\{1,\ldots,K\!-\!2\}$), followed by the blank tokens $(0)$ for $L$ steps, and a ``go-cue'' symbol $(K\!-\!1)$, indicating the beginning of the output timestep. The remaining $S-1$ steps are again the blank tokens $(0)$. The target sequence is blank for the first $S+L$ steps and then reproduces the original $S$ symbols in order over the final $S$ steps, immediately upon the go-cue symbol.

We used a vanilla RNN with $H=128$ hidden units and a $\tanh$ nonlinearity. Models were trained for 10 epochs with batch size 256 using SGD with momentum 0.9 and learning rate $10^{-4}$ for all WP methods and $10^{-3}$ for backprop (all WP methods could not learn with learning rate $10^{-3}$, likely due to the large variance in the estimated gradient). The recurrent weights ($W_{hh}$), input weights ($W_{xh}$), and hidden bias ($b_h$) were updated using weight perturbation, while the readout weights ($W_{hy}$) and bias ($b_y$) were updated using exact gradients. The norm of the weight perturbation was rescaled to $\epsilon_{WP}\cdot \sqrt{NM}$, where $N$ and $M$ is the number of rows and columns of the weight matrix and $\epsilon_{WP}=10^{-4}$, across different WP methods. The manifold used for perturbations was estimated online by incremental PCA on hidden states (32 PCs), updated every batch.

\subsection{Data Preprocessing}
\label{app:data_preprocessing}

\paragraph{CIFAR-10.}
Images are normalized using channel-wise mean and standard deviation computed from the training set:
\begin{align}
    \text{mean} &= [0.4914, 0.4822, 0.4465] \\
    \text{std} &= [0.2470, 0.2435, 0.2616]
\end{align}
No data augmentation is applied during training.

\paragraph{ImageNet.}
We use standard ImageNet preprocessing:

\textit{Training augmentation:}
\begin{enumerate}
    \item \texttt{RandomResizedCrop(224)}: Random crop with scale $(0.08, 1.0)$ and aspect ratio $(3/4, 4/3)$, resized to $224 \times 224$
    \item \texttt{RandomHorizontalFlip()}: Horizontal flip with probability 0.5
    \item \texttt{ToTensor()}: Convert to tensor and scale to $[0, 1]$
    \item Normalization: mean $= [0.485, 0.456, 0.406]$, std $= [0.229, 0.224, 0.225]$
\end{enumerate}

\textit{Test preprocessing:}
\begin{enumerate}
    \item \texttt{Resize(256)}: Resize shorter edge to 256 pixels
    \item \texttt{CenterCrop(224)}: Center crop to $224 \times 224$
    \item \texttt{ToTensor()}: Convert to tensor
    \item Same normalization as training
\end{enumerate}

%%%%%%%%%%%%%%%%%%%%%%%%%%%%%%%%%%%%%%%%%%%%%%%%%%%%%%%%%%%%%
\section{Algorithm Implementation Details}
\label{app:implementation}

\subsection{Feedback Weight Initialization}
\label{app:init}

The feedback weights $B_l \in \mathbb{R}^{n_o \times n_l}$ (where $n_o$ is the number of output units and $n_l$ is the flat dimension of layer $l$'s activations) can be initialized in two ways:

\paragraph{Initial Jacobian (InitJac).}
We compute the Jacobian $\frac{\partial \mathbf{y}}{\partial \mathbf{x}_l}$ at network initialization by:
\begin{enumerate}
    \item Sampling a small batch of random inputs (batch size 32)
    \item Computing the Jacobian using PyTorch's \texttt{torch.func.jacrev} and \texttt{torch.func.vmap} for vectorized computation
    \item Averaging over the batch to obtain $B_l^{(0)}$
\end{enumerate}

% To avoid GPU memory overflow, the Jacobian is computed in chunks:
% \begin{itemize}
%     \item Layers 0--1 (conv1--2): chunk size 4
%     \item Layers 2--4 (conv3--5): chunk size 8
%     \item Layers 5--6 (fc1--2): chunk size 16
% \end{itemize}
% We apply aggressive memory clearing between chunks (\texttt{torch.cuda.empty\_cache()} and \texttt{gc.collect()}).

\paragraph{Random Initialization (No InitJac).}
For comparison, we also test random initialization where the elements of the initial Jacobian are randomly permuted, destroying any meaningful gradient structure while preserving the overall statistics.

\subsection{Incremental PCA Implementation}
\label{app:ipca}

We implement incremental PCA algorithm \cite{ross_incremental_2008} in PyTorch for GPU acceleration. It maintains running estimates of the feature-wise mean and variance as well as the top-$k$ principal axes for $d$-dimensional activations. Each update is performed on a minibatch $X\in\mathbb{R}^{b\times d}$ (requiring $b\ge k$). The batch is centered and combined with the previous decomposition by vertically stacking (i) the prior components scaled by their singular values, (ii) the centered current batch, and (iii) a mean-correction term that accounts for changes in the running mean. An SVD of this augmented matrix yields updated principal axes and singular values; we apply the standard SVD sign-flip convention for deterministic component orientations. The algorithm returns the current principal axes as a $d\times k$ matrix (and returns random unit-norm vectors before the first update).

The SVD step can optionally use a randomized low-rank routine (\texttt{torch.svd\_lowrank}); in all experiments reported here we instead compute the exact SVD via \texttt{torch.linalg.svd}, which we found faster in our setting. Finally, to keep PCA updates asynchronous without stalling training, activation minibatches are streamed to each incremental PCA instance through a bounded queue whose capacity is set by \texttt{max\_queue\_batches}; if this queue is full, new minibatches are dropped rather than blocking the main loop, bounding memory usage and limiting staleness of PCA updates. However, in all experiments reported here, we made sure that PCA updates were fast enough to keep up with model training without accumulating activation minibatches.

\subsection{Multiprocessing for incremental PCA}

For CIFAR-10 and ImageNet experiments, NMNC training used a two-GPU multiprocessing setup to parallelize model training and incremental PCA updates:

\begin{itemize}
    \item \textbf{Main process (GPU 0)}: Runs forward and backward passes, performs weight updates, and coordinates training.
    \item \textbf{PCA workers (GPU 1)}: Separate processes for each hidden layer (Conv1--3 and FC1 for CIFAR-10; Conv1--5 and FC1--2 for ImageNet), each maintaining its own incremental PCA state.
    \item \textbf{Communication}:
    \begin{itemize}
        \item Activations are sent from the main process to PCA workers via per-layer multiprocessing queues.
        \item Updated principal components are returned to the main process via a shared result queue.
        \item PyTorch CUDA inter-process communication (IPC) is used for efficient GPU tensor transfer.
    \end{itemize}
    \item \textbf{Non-blocking execution}: PCA updates run asynchronously with training; the main loop proceeds while workers process accumulated activations.
\end{itemize}

%%%%%%%%%%%%%%%%%%%%%%%%%%%%%%%%%%%%%%%%%%%%%%%%%%%%%%%%%%%%%
\section{Intrinsic Dimensionality Analysis}
\label{app:intrinsic_dim}

\subsection{Network Width Scaling}

To study how manifold dimensionality scales with network size, we vary the width multiplier $n$ applied to all channel dimensions:

\begin{table}[h]
\centering
\caption{Network configurations for width scaling experiments.}
\begin{tabular}{@{}llllll@{}}
\toprule
Multiplier $n$ & conv1 & conv2 & conv3 & fc1 & Total params (approx.) \\
\midrule
1/32 & 2 & 4 & 8 & 32 & 3K \\
1/16 & 4 & 8 & 16 & 64 & 12K \\
1/8 & 8 & 16 & 32 & 128 & 47K \\
1/4 & 16 & 32 & 64 & 256 & 186K \\
1/2 & 32 & 64 & 128 & 512 & 740K \\
1 (default) & 64 & 128 & 256 & 1024 & 2.9M \\
2 & 128 & 256 & 512 & 2048 & 11.7M \\
4 & 256 & 512 & 1024 & 4096 & 46.7M \\
8 & 512 & 1024 & 2048 & 8192 & 186M \\
\bottomrule
\end{tabular}
\end{table}

\section{Why low-rank perturbations do not change the fundamental variance scaling of weight perturbation}
\label{app:lowrank_wp_variance}

This appendix provides a short calculation supporting the statement in Footnote~4 (Section~4.5):
\emph{using low-rank perturbations can reduce the \textbf{computational} cost of each perturbation, but it does
not fundamentally remove the scaling of the \textbf{gradient-estimator variance} with the number of parameters
being perturbed.} The key point is that unless we restrict optimization to a lower-dimensional
\emph{parameterization} (e.g., only optimizing a low-rank factorization), the gradient being estimated still lives
in a $d$-dimensional space, where $d$ is the number of perturbed parameters.

\paragraph{Setup (matrix weight perturbation).}
Let $W \in \mathbb{R}^{N\times M}$ be a weight matrix (e.g., $W_{hh}$ in the RNN experiments), and let
$\mathcal{L}(W)$ be the scalar loss. Denote the true gradient by
\[
G \;:=\; \nabla_W \mathcal{L}(W) \in \mathbb{R}^{N\times M},
\qquad d := NM.
\]
A common (antithetic) weight-perturbation / ES estimator uses a random perturbation $E \in \mathbb{R}^{N\times M}$
and
\begin{equation}
\hat G(E)
\;:=\;
\frac{\mathcal{L}(W+\varepsilon E)-\mathcal{L}(W-\varepsilon E)}{2\varepsilon}\,E,
\label{eq:wp_estimator}
\end{equation}
where $\varepsilon>0$ controls perturbation magnitude. For $K$ i.i.d.\ perturbations $\{E_k\}_{k=1}^K$,
we average $\hat G_K := \frac{1}{K}\sum_{k=1}^K \hat G(E_k)$.

\paragraph{Small-$\varepsilon$ approximation.}
For sufficiently small $\varepsilon$, a first-order Taylor expansion gives
\[
\mathcal{L}(W\pm \varepsilon E) \;=\; \mathcal{L}(W) \pm \varepsilon \langle G, E\rangle + O(\varepsilon^2),
\]
where $\langle A,B\rangle := \mathrm{tr}(A^\top B)$ is the Frobenius inner product. Substituting into
\eqref{eq:wp_estimator} yields
\begin{equation}
\hat G(E) \;=\; \langle G, E\rangle E + O(\varepsilon^2).
\label{eq:wp_linearized}
\end{equation}

\paragraph{Isotropy and unbiasedness.}
Write $e := \mathrm{vec}(E)\in\mathbb{R}^d$ and $g := \mathrm{vec}(G)\in\mathbb{R}^d$.
If the perturbations are (second-moment) isotropic,
\begin{equation}
\mathbb{E}[e] = 0,
\qquad
\mathbb{E}[e e^\top] = I_d,
\label{eq:isotropy}
\end{equation}
then, ignoring $O(\varepsilon^2)$ terms, \eqref{eq:wp_linearized} is unbiased:
\[
\mathbb{E}[\hat G(E)] \;=\; \mathbb{E}[\langle G,E\rangle E] \;=\; G.
\]
Importantly, the normalization used in low-rank schemes is typically chosen precisely so that
\eqref{eq:isotropy} (or a scaled version) holds.

\paragraph{Variance as Frobenius MSE scales with $d$.}
A natural global measure of estimator noise is the Frobenius mean-squared error (MSE)
\[
\mathbb{E}\big[\|\hat G_K - G\|_F^2\big]
\;=\;
\frac{1}{K}\,\mathbb{E}\big[\|\hat G(E)-G\|_F^2\big]
\quad (\text{i.i.d.\ samples}).
\]
Using \eqref{eq:wp_linearized} and unbiasedness,
\begin{equation}
\mathbb{E}\big[\|\hat G(E)-G\|_F^2\big]
\;=\;
\mathbb{E}\big[\langle G,E\rangle^2\|E\|_F^2\big] - \|G\|_F^2
\;+\; O(\varepsilon^2).
\label{eq:mse_general}
\end{equation}
For many isotropic choices (including full i.i.d.\ Gaussian perturbations and normalized low-rank
perturbations), the leading term in \eqref{eq:mse_general} grows linearly with $d=NM$.

\subparagraph{Full-rank i.i.d.\ Gaussian perturbations.}
If $e\sim\mathcal{N}(0,I_d)$ (equivalently $E_{ij}\stackrel{i.i.d.}{\sim}\mathcal{N}(0,1)$), standard Gaussian
fourth-moment identities imply
\[
\mathbb{E}\big[\langle G,E\rangle^2\|E\|_F^2\big] \;=\; (d+2)\|G\|_F^2,
\]
and therefore
\begin{equation}
\mathbb{E}\big[\|\hat G_K - G\|_F^2\big]
\;=\;
\frac{d+1}{K}\,\|G\|_F^2
\;+\; O(\varepsilon^2).
\label{eq:mse_gaussian}
\end{equation}
Thus, to keep the \emph{global} estimator noise (in Frobenius norm) constant as $d$ grows, one needs
$K=\Omega(d)$ perturbation samples.

\subparagraph{Rank-1 perturbations (explicit calculation).}
Consider rank-1 perturbations
\begin{equation}
E = u v^\top,
\qquad
u\sim\mathcal{N}(0,I_N),\;\; v\sim\mathcal{N}(0,I_M),
\label{eq:rank1}
\end{equation}
which correspond to $e = v\otimes u$ in vectorized form. One can verify that
$\mathbb{E}[e e^\top] = I_M\otimes I_N = I_d$, so \eqref{eq:isotropy} holds and the estimator is unbiased
(up to $O(\varepsilon^2)$).

In this case $\langle G,E\rangle = u^\top G v$ and $\|E\|_F^2 = \|u\|^2\|v\|^2$.
A direct Gaussian-moment calculation gives
\[
\mathbb{E}\big[(u^\top G v)^2 \,\|u\|^2\|v\|^2\big]
\;=\;
(N+2)(M+2)\,\|G\|_F^2,
\]
and plugging into \eqref{eq:mse_general} yields
\begin{equation}
\mathbb{E}\big[\|\hat G_K - G\|_F^2\big]
\;=\;
\frac{(N+2)(M+2)-1}{K}\,\|G\|_F^2
\;+\; O(\varepsilon^2)
\;=\;
\frac{d + 2N + 2M + 3}{K}\,\|G\|_F^2
\;+\; O(\varepsilon^2).
\label{eq:mse_rank1}
\end{equation}
The leading term is still $\Theta(d/K)$, i.e., the same \emph{dimension-driven} scaling as the full-rank
Gaussian case \eqref{eq:mse_gaussian}, up to constants.

\paragraph{Rank-$r$ low-rank perturbations.}
A common rank-$r$ construction is
\begin{equation}
E
=
\frac{1}{\sqrt{r}}\sum_{k=1}^r u_k v_k^\top,
\qquad
u_k\sim\mathcal{N}(0,I_N),\;\; v_k\sim\mathcal{N}(0,I_M)\;\; \text{i.i.d.}
\label{eq:rankr}
\end{equation}
The $1/\sqrt{r}$ normalization ensures $\mathbb{E}[e e^\top]=I_d$ (each entry has $O(1)$ variance), so the
estimator remains unbiased to first order. Increasing $r$ changes higher-order moments (and thus the
\emph{constant} in the MSE), but as long as we are still estimating a $d$-dimensional gradient over the
full parameter space, the leading MSE scaling remains proportional to $d/K$.

\paragraph{Interpretation.}
Low-rank perturbations can be valuable because applying $E$ (or generating it) may cost
$O(r(N+M))$ rather than $O(NM)$, improving computational/hardware efficiency. However, unless learning is
\emph{restricted to a lower-dimensional parameterization} (so the \emph{unknown} gradient itself has only
$O(r(N+M))$ degrees of freedom), the Monte Carlo estimator is still recovering a $d$-dimensional object.
Consequently, the number of perturbations required to control the \emph{global} estimator noise scales as
$K=\Omega(d)$, and low-rank perturbations primarily affect constants rather than this fundamental scaling.

\section{TwoNN intrinsic dimension estimator}
\label{app:twonn}

We estimate intrinsic dimensionality using the TwoNN estimator of \citet{facco_estimating_2017}, which uses only the first and second nearest neighbors of each point. Given a dataset $\{x_i\}_{i=1}^N \subset \mathbb{R}^D$, let $r_{i,1}$ and $r_{i,2}$ denote the Euclidean distances from $x_i$ to its first and second nearest neighbors (excluding $x_i$), and define the ratio
\[
\mu_i := \frac{r_{i,2}}{r_{i,1}} \in [1,\infty).
\]
Under the assumption that, within the scale set by $r_{i,2}$, the sampling density is approximately constant on a locally $d$-dimensional manifold, the cumulative distribution of $\mu$ depends only on $d$ (and not on the density):
\[
F(\mu)=\Pr(\mu_i \le \mu) = 1 - \mu^{-d}, \qquad \mu \ge 1.
\]
This implies the linear relation
\[
-\log\!\bigl(1-F(\mu)\bigr) = d\,\log \mu,
\]
so the points $(\log \mu_{(i)}, -\log(1-\hat F(\mu_{(i)})))$ lie approximately on a line through the origin with slope $d$, where $\mu_{(i)}$ are the sorted ratios and $\hat F(\mu_{(i)})=(i-0.5)/N$ is the empirical CDF. We estimate $\hat d$ by least-squares fitting of this line; we discard the top and bottom 10\% of $\mu_{(i)}$ when fitting to reduce sensitivity to outliers and boundary effects.

\newpage
\section{Supplementary figures}
\label{app:sup_fig}

\begin{figure}[h]
\centering
\includegraphics[width=\linewidth]{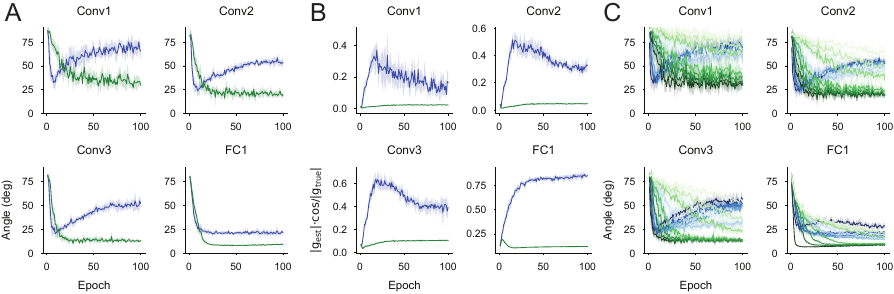}
\caption{Same as \Cref{fig:4} for alignment in weight space.}
\label{fig:8}
\end{figure}

\begin{figure}[h]
\centering
\includegraphics[width=\linewidth]{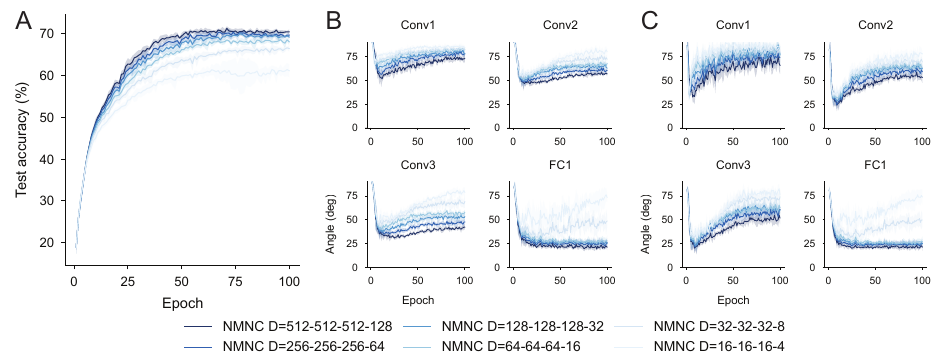}
\caption{NMNC with varying numbers of neural manifold dimensions (PCs) for the CIFAR-10 model. (A) Test accuracy vs. epochs. (B) Alignment between true and estimated gradients in activation space. (C) Alignment in weight space.}
\label{fig:9}
\end{figure}

\newpage
\begin{figure}[h]
\centering
\includegraphics[width=\linewidth]{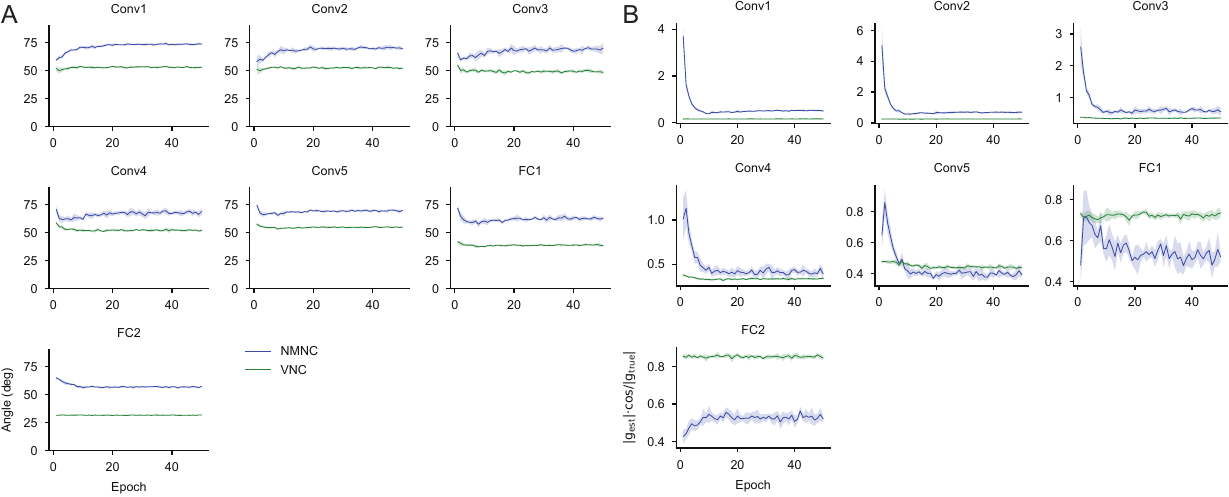}
\caption{Alignment between true and estimated gradients in activation space for the ImageNet model. (A) Cosine similarity angle between the true and estimated gradients for NMNC and VNC across layers. (B) Normalized magnitude of the estimated gradient projected onto the true
gradient direction for NMNC and VNC across layers.}
\label{fig:10}
\end{figure}

\begin{figure}[h]
\centering
\includegraphics[width=\linewidth]{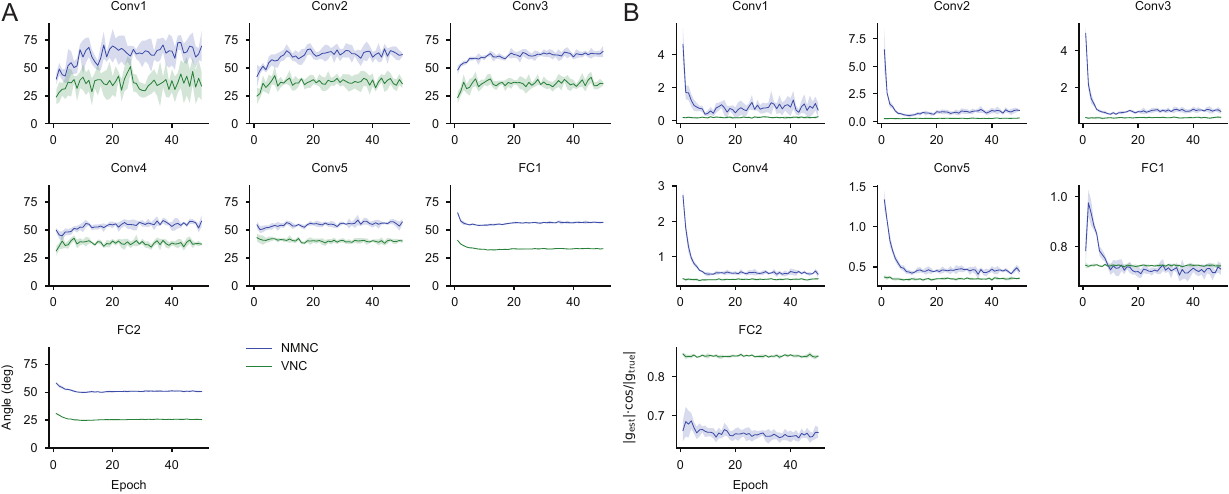}
\caption{Same as \Cref{fig:10} for alignment in weight space.}
\label{fig:11}
\end{figure}
%%%%%%%%%%%%%%%%%%%%%%%%%%%%%%%%%%%%%%%%%%%%%%%%%%%%%%%%%%%%%%%%%%%%%%%%%%%%%%%
%%%%%%%%%%%%%%%%%%%%%%%%%%%%%%%%%%%%%%%%%%%%%%%%%%%%%%%%%%%%%%%%%%%%%%%%%%%%%%%

\end{document}